\documentclass[runningheads, envcountsame, a4paper]{llncs}
\usepackage{latexsym}
\usepackage{amsfonts}
\usepackage{amsmath}
\usepackage{multirow}
\usepackage[hyphens]{url}
\usepackage{microtype}
\usepackage{graphicx}

\begin{document}

\title{Mutual Information Decay Curves and Hyper-Parameter Grid Search Design for Recurrent Neural Architectures}
\titlerunning{Mutual Information \& Recurrent Neural Architectures}
\author{Abhijit~Mahalunkar\orcidID{0000-0001-5795-8728} \and John~D.~Kelleher\orcidID{0000-0001-6462-3248}}
\authorrunning{A.~Mahalunkar and J.~D.~Kelleher}
\institute{ADAPT Research Center, Technological University Dublin, Ireland \\
\email{$\{$abhijit.mahalunkar,john.d.kelleher$\}$@tudublin.ie}}
\toctitle
\tocauthor
\maketitle
\setcounter{footnote}{0}

\begin{abstract}
We present an approach to design the grid searches for hyper-parameter optimization for recurrent neural architectures. The basis for this approach is the use of mutual information to analyze long distance dependencies (LDDs) within a dataset. We also report a set of experiments that demonstrate how using this approach, we obtain state-of-the-art results for DilatedRNNs across a range of benchmark datasets.
\end{abstract}

\section{Introduction}
Recurrent neural networks trained using backpropagation through time suffer from exploding or vanishing gradients~\cite{Hochreiter1991,Hochreiter01gradientflow,Yoshua1994,kelleher2019}. This problem presents a specific challenge in modeling sequential datasets which exhibit long distance dependencies (LDDs)~\cite{Mahalunkar2018}. LDDs describe an interaction between two (or more) elements in a sequence that are separated by an arbitrary number of positions. This results in the decay of statistical dependence of two points with increasing distance between them. Building recurrent neural architectures that are able to model LDDs is an open research problem and much of the current research on the topic focuses on designing new neural architectures. Early work in this direction proposed a hierarchical recurrent neural network that introduced several levels of state variables, working at different time scales~\cite{ElHihi1995}. This work inspired other architectures, such as DilatedRNN~\cite{chang2017dilated}, Skip RNN~\cite{campos2018skip}, etc. Other well-known approaches to address this challenge are~\cite{hochreiter1997lstm,Merity2016,Graves2014,Salton2017,vaswani_2017,Dai2019}. In this paper, we argue that a key step in designing recurrent neural architectures is to understand the decay of dependence in sequential data and to use this understanding to inform the setting of the relevant hyper-parameters of the architecture. In previous work, we developed an algorithm to compute and visualize the decay of dependence of the symbols within the dataset~\cite{2018bMahalunkar}. In this paper, we build on this previous work and show how this type of analysis can inform the selection of hyper-parameters of existing recurrent neural architectures. In this regard, we use DilatedRNNs~\cite{chang2017dilated} as a test model, and for several benchmark datasets, we study the decay of dependence of a dataset and then design a set of dilations tailored to that dataset. With this approach, we achieve better performance as compared to the technique mentioned in the original implementation~\cite{chang2017dilated}.

\section{Dilated Recurrent Neural Networks}\label{sec:dilatedrnns}
In this paper, we demonstrate how the analysis of the decay of dependence is a useful source of information to guide the selection of hyper-parameters to model the sequential data. We have chosen to DilatedRNNs~\cite{chang2017dilated} as test model for our experiments because of the relatively transparent relationship between some of the hyper-parameters of this architecture and the ability of the network to model LDDs. The DilatedRNN architecture is a multi-layer and cell-independent architecture characterized by multi-resolution dilated recurrent skip-connections. This alleviates the gradient problems and extends the range of temporal dependencies. Upon stacking multiple dilated recurrent layers with increasing skip-connections, these networks can learn temporal dependencies at different scales. DilatedRNN architecture and the dilations are described in~\cite{chang2017dilated}. The size of dilations per layer and the number of layers are supplied using the \emph{dilations} hyper-parameter. This hyper-parameter controls the gradient flow and memory capacity of the DilatedRNNs. We hypothesize that to optimize the performance of a DilatedRNN on a dataset, the set of dilations should be tailored to match LDDs within the dataset. In particular: 1) the max dilation should match the span of LDDs present in the dataset (i.e. the dilation length should not extend to distances where there is low mutual information), 2) the increase in the size of dilation per layer should match the decay of dependence observed in the dataset (i.e. in regions where there is a rapid decay of dependence, there should be dense set of skip-connections capturing the dependence).

The process we propose for fitting the dilations hyper-parameter of these networks to a dataset is as follows: 1) plot the dependency decay curve~\cite{2018bMahalunkar}, 2) design a grid search over the hyper-parameter space using the framework presented in this paper, 3) fit the model to the data and evaluate. To demonstrate this approach, we train DilatedRNNs on MNIST and Penn Treebank (PTB).

\section{Interpreting Dependency Decay Curve to Inform Hyper-Parameter Optimization}\label{sec:interpreting}
\begin{figure}
\centering
\includegraphics[scale=0.2]{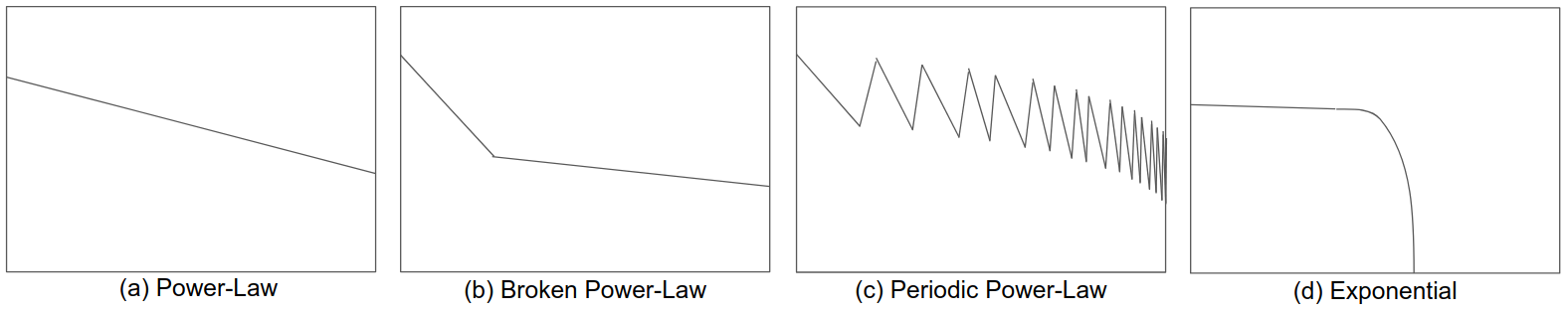}
\caption{Dependency decay curve framework which displays standard LDD curves}
\label{fig:framework}
\end{figure}

In~\cite{2018bMahalunkar}, we proposed an algorithm to analyze the decay of dependence in sequential datasets. This analysis can be visualised by plotting this decay on a log-log axis, where the x-axis describes the distance $d$ between two observations in the sequence and the y-axis describes the mutual information (MI) at that $d$ (in \emph{nats}). We call these plots dependency decay curves and this paper focuses on how the analysis of these plots can be used to guide hyper-parameter selection. The decay of dependence either follows exponential decay, indicating the absence of LDDs, or power-law decay, indicating the presence of LDDs. Furthermore, the influence of various phenomena on LDDs can lead to decay curve following: 1) power-law decay, 2) power-law decay with a periodicity present, or 3) broken power-law. A broken power-law is made up of multiple power-laws joined at inflection point. Fig.~\ref{fig:framework} illustrates how these different types of decay are exhibited in an dependency decay curve plot. As dependency decay curves are plotted on a log-log axis a curve that follows a straight line represents a power-law decay (e.g. Fig.~\ref{fig:framework}(a)).

\begin{figure}
\centering
\includegraphics[scale=0.2]{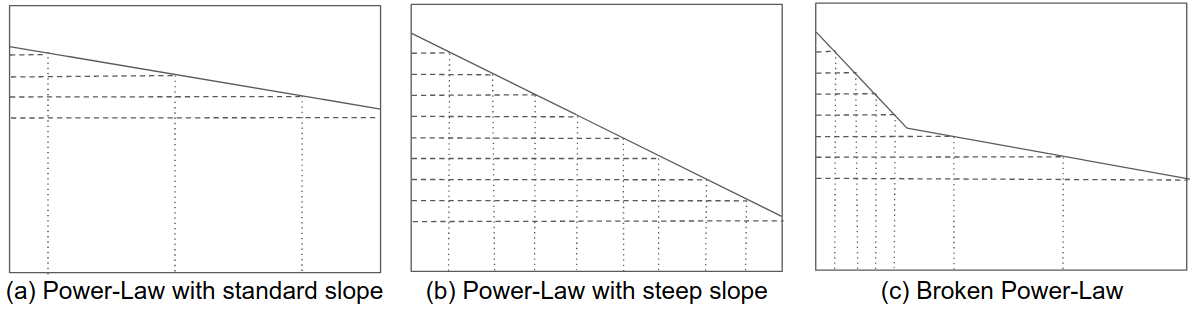}
\caption{Mapping set of dilations to dependency decay curve}
\label{fig:dilations_power_law}
\end{figure}

In the context of DilatedRNNs the best dilation hyper-parameters to model a dataset exhibiting power-law decay are: 1) max dilation should be equal to $d$ where the dependency decay curve crosses MI${=}10^{-5}$ (this threshold is informed by~\cite{Lin2017} as MI below this is assumed to be noise); and 2) the set of dilations should be designed such that the density of skip-connections in a region should increase when the slope is high and decrease when it is low. The standard progression of dilations used for DilatedRNNs is $1,2,4,8,\dots$ and these dilations are a good guide for reasonable rates of dependency decay (similar to the slope in Fig.~\ref{fig:framework}(a)). However, if the line gets steeper the spacing between the skip-connections should become smaller and denser e.g. $1,2,3,4,5,\dots$. Fig.~\ref{fig:dilations_power_law} illustrates the design pattern we use to connect the slope of the decay of dependence to the density of the skip-connections in a DilatedRNN. In Fig.~\ref{fig:dilations_power_law}(a) and~\ref{fig:dilations_power_law}(b), we see two power-law decays in two plots. In each plot we have drawn a sequence of equidistant horizontal lines representing hidden layers. We use the density of the $x$-intercept of every horizontal line to inform the patterning of the skip-connections for every layer in the network.

Fig.~\ref{fig:framework}(b) is an dependency decay curve plot that exhibits a broken power-law~\cite{Rhoads_1999}. The hyper-parameter selection for such plots follow the same rules as that of the power-law decay curve, with one major difference being that the presence of multiple power-laws will require different patterning of the dilations in different segments of the broken power-law in order to model the different rates of MI decay. Fig.~\ref{fig:dilations_power_law}(c), shows that when the steeper power-law changes to a shallower power-law, the dilations become more sparse. Fig.~\ref{fig:framework}(c) is a dependency decay curve plot that is exhibiting a power-law decay with periodicity. The presence of periodicity within the decay indicates that the max dilation parameter should be set to the period of the MI peaks. The set of dilations can still follow the standard pattern for DilatedRNNs. An exponential decay involves a rapid decay in mutual information, such as that shown in Fig.~\ref{fig:framework}(d). Such a rapid decay indicates the absence of LDDs. So in order to model such datasets, the max dilation should be set to where the dependency decay curve crosses the MI${=}10^{-5}$. It can follow the standard set of dilations of the DilatedRNNs. 

\section{Dependency Decay Curves of Benchmark Datasets}\label{sec:plot_ldds}
\subsection{Sequential MNIST}
\begin{figure}
  \centering
  \begin{tabular}[c]{cc}
    \includegraphics[scale=0.32]{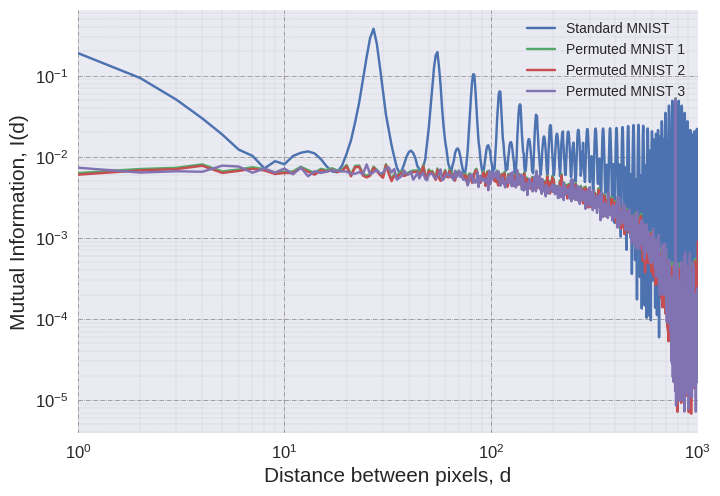}
    &
    \includegraphics[scale=0.32]{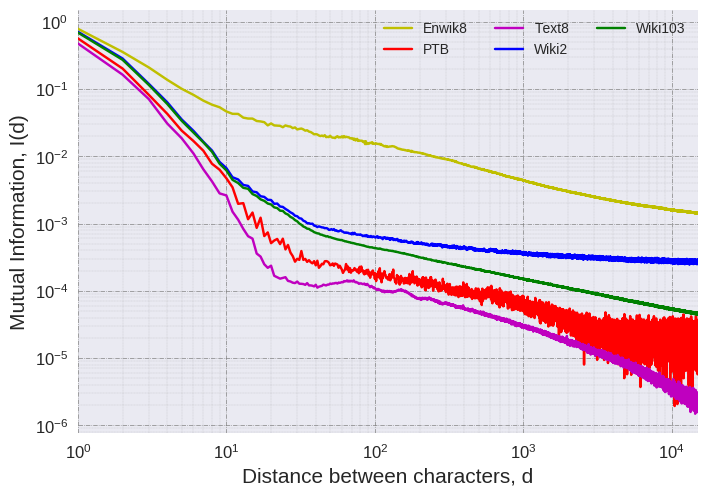} \\
    (a) Unpermuted \& Permuted sequential & (b) Character-based language \\
    MNIST (multiple seeds) &
  \end{tabular}
  \caption{Dependency Decay Curves}
  \label{fig:mnist_lm_chars}
\end{figure}

Sequential MNIST is widely used to evaluate recurrent neural architectures. It contains $240000$ training and $40000$ test images. Each of these is $28$x$28$ pixels in size, and each pixel can take one of $256$ pixel values. In order to use them in a sequential task, the $2$D images are converted into a $1$D vector of $784$ pixels by concatenating all the rows of the pixels. This transformation results in pixel dependencies which span up to approximately $28$ pixels. These dependencies arise due to high correlation of a pixel with its neighboring pixels. The structure of the Sequential MNIST dataset is such that its dependency decay curve is likely to contain regular peaks and troughs. We plot the dependency decay curve of the unpermuted and permuted sequential MNIST datasets in Fig.~\ref{fig:mnist_lm_chars}(a). 

Standard sequential MNIST exhibits high MI at $D{=}1$ indicating strong dependencies at close proximity. The dependencies then decay as a function of power-law. Hence, in-order to fully capture these dependencies, the recurrent neural architectures should maintain gradients/attention across multiple timescales as a function of power-law to accurately model these dependencies. However, we see peaks of MI at intervals of $28$ due to pixel dependencies. The regular peaks in the decay curve indicate that the span of the dependencies lie within $D{\approx}28$.

We generated \emph{permuted} versions of the sequential MNIST dataset with multiple seeds for use as a comparator with the unpermuted sequential MNIST. When we examine the dependency decay curves of permuted MNIST datasets, we observe that the dependencies are substantially less between close-by symbols (pixels in this case), e.g. for $D{=}1$ the green, red and purple lines are much lower than the blue line. This is a result of permutations applied to the data which disrupt spatial dependencies. Another impact of this disruption is the relatively flat curve for $D{<}300$ which indicates an absence of spatial dependencies. In-order to model these datasets that exhibit a relatively flat curve, the recurrent neural architectures requires uniform distribution of attention/gradients across all time scales. However, beyond $D{>}300$, we observe exponential decay of dependence, where the value of MI falls below $10^{-5}$ indicating no further dependencies. This point ($D{\approx}780$) indicates the span of the dependencies and the limit on the memory capacity of recurrent neural architectures.

\subsection{Character-Based Datasets}\label{ssec:char_ldds}
There are a number of natural language datasets that are frequently used to benchmark recurrent neural architectures, such as the Penn TreeBank (PTB), Wiki2, Wiki103, Text8 and Enwik8. These datasets can either be analysed at the word or character level. In this section we focus our analysis at the character level. Fig.~\ref{fig:mnist_lm_chars}(b) shows the dependency decay curve of these character-based datasets. These plots follow a power-law decay function. Except Enwik8, which follows a single power-law decay, all the other datasets (PTB, Wiki2, Wiki103 and Text8) follow multiple \emph{power-laws} with an inflection point\footnote{Enwik8 exhibits a single power-law due to the presence of XML code (strict markup grammar and long contextual dependencies). Consequently, the dependency decay curve of Enwik8 are different from the rest of the character-based datasets.}. The character-based datasets exhibit two distinct decay curves i) single power-law decay (Enwik8), and ii) broken power-law decay (the rest of the curves). The optimal hyper-parameters are selected based on the framework discussed in section~\ref{sec:interpreting}.

\section{Optimising Hyper-parameters of Recurrent Neural Architectures}\label{sec:drnn}
\subsection{Experiments with Sequential MNIST}
In this experiment, we trained DilatedRNNs with unpermuted and permuted sequential MNIST datasets in a classification task (classify digits $0$-$9$ from their images). The original paper that introduced DilatedRNNs~\cite{chang2017dilated} used the same max dilation hyper-parameter for both of these datasets i.e. $256$, and a standard set of dilations (i.e., $1,2,4,8,\dots$). The best results reported by~\cite{chang2017dilated} for these two datasets were: unpermuted sequential MNIST $99.0$/$99.2$ and permuted sequential MNIST $95.5$/$96.1$. However, our analysis of these datasets has revealed different max dependencies across these dataset. For unpermuted sequential MNIST we identified a periodicity of $28$ and so we expected the max dilation value to be near $28$ to deliver better performance. In permuted sequential MNIST we identified that the dependencies extend up to $780$ and so we would expect better performance by extending the max dilation up to this value.
\begin{table}
\small
\centering
\caption{Results for sequential MNIST using GRU cells}
\label{tab:gru_unper}
\begin{tabular}{cccc}
\hline
\# of & Set of & Hidden & Accuracy \\
 Layers & Dilations & per Layer & \\
\hline
$4$ & $1,2,4,8$ & $20/50$ & $98.96/99.18$ \\
$5$ & $1,2,4,8,16$ & $20/50$ & $98.94/99.21$ \\
$6$ & $1,2,4,8,16,32$ & $20/50$ & $\mathbf{99.17/99.27}$ \\
$7$ & $1,2,4,8,16,32,64$ & $20/50$ & $99.05/99.25$ \\
$8$ & $1,2,4,8,16,32,64,128$ & $20/50$ & $99.15/99.23$ \\
$9$ & $1,2,4,8,16,32,64,128,256$ & $20/50$ & $98.96/99.17$ \\
\hline
\end{tabular}
\end{table}
\begin{table}
\small
\centering
\caption{Results for permuted sequential MNIST using RNN cells}
\label{tab:rnn_per}
\begin{tabular}{cccc}
\hline
\# of & Set of & Hidden & Accuracy \\
 Layers & Dilations & per Layer & \\
\hline
$7$ & $1,2,4,8,16,32,64$ & $20/50$ & $95.04/95.94$ \\
$8$ & $1,2,4,8,16,32,64,128$ & $20/50$ & $95.45/95.88$ \\
$9$ & $1,2,4,8,16,32,64,128,256$ & $20/50$ & $95.5/96.16$ \\
$10$ & $1,2,4,8,16,32,64,128,256,512$ & $20/50$ & $95.62/96.4$ \\
$11$ & $1,2,4,8,16,32,64,128,256,512,780$ & $20/50$ & $\mathbf{95.66/96.47}$ \\
\hline
\end{tabular}
\end{table}

To test these hypotheses we trained DilatedRNNs with various sets of dilations. To keep our results comparable with those reported in~\cite{chang2017dilated} the original code\footnote{\url{https://github.com/code-terminator/DilatedRNN}} was kept unchanged except for the max dilation hyper-parameter. The test results of these experiments are in tables~\ref{tab:gru_unper} and~\ref{tab:rnn_per}. For unpermuted task, the model delivered best performance for max dilation of $32$. Focusing on the results of the permuted sequential MNIST, the best performance was delivered with the max dilation of $780$. These results confirm that the best performance is obtained when the max dilation is similar to the span of the LDDs of a given dataset.

\subsection{Experiments with Character-Based Datasets}
In this experiment, we trained DilatedRNN with PTB dataset in a language modeling task. The standard evaluation metric of language models is perplexity and it is a measure of the confusion of the model when making predictions. The original DilatedRNNs paper~\cite{chang2017dilated} reported best performance on PTB with a perplexity of $1.27$ using GRU cells and max dilation of $64$. However, the original DilatedRNN implementations used for this task has not been released. Hence we used another implementation of DilatedRNN\footnote{\url{https://github.com/zalandoresearch/pytorch-dilated-rnn}}. Having done multiple experiments with this new implementation, including using the same hyper-parameters as those reported in~\cite{chang2017dilated}, we found that the perplexity was always higher than the original implementation. We attribute this to the fact that~\cite{chang2017dilated} does not report all the hyper-parameter settings they used and so we had to assume some of the hyper-parameters (other than the dilations) used for this task. The test perplexity results of these experiments are in table~\ref{tab:gru_ptb}.
\begin{table}
\small
\centering
\caption{Results for character-based datasets (PTB) using GRU cells}
\label{tab:gru_ptb}
\begin{tabular}{cccccc}
\hline
Model & \# of & Set of & Hidden & \# of & Test Perplexity \\
\# & Layers & Dilations & per Layer & Parameters & (bpc)\\
\hline
$1.$ & $7$ & $1,2,4,8,16,32,64$ & $256$ & $2660949$ & $1.446$ \\
$2.$ & $8$ & $1,2,4,8,16,32,64,128$ & $256$ & $3055701$ & $1.471$ \\
$3.$ & $12$ & $1,2,4,8,16,32,64,128,256,\dots$ & $256$ & $4634709$ & $1.561$ \\
&& $512,1024,2048$ &&&\\
$4.$ & $7$ & $1,2,3,4,5,6,8$ & $256$ & $2660949$ & $1.416$ \\
$5.$ & $8$ & $1,2,4,8,16,32,73,240$ & $256$ & $3055701$ & $1.456$ \\
$6.$ & $12$ & $1,2,3,4,5,6,8,11,18,32,73,240$ & $256$ & $4634709$ & $\mathbf{1.386}$ \\
\hline
\end{tabular}
\end{table}

Recall from section~\ref{ssec:char_ldds}, that the character-based PTB dataset exhibits a broken power-law dependency decay. Consequently, following the framework we presented, we expect that a model that has set of dilations that follow the pattern that we have proposed for broken power-laws (with different densities of connections on either side of the inflection point) should outperform a model that uses a standard set of dilations. This experiment was designed to test this hypothesis while controlling the model size. We designed six different model architectures (listed in table~\ref{tab:gru_ptb}), three using the standard dilations (model\# $1,2,$ and $3$) and three using dilation patterns informed by the dependency decay curve analysis of PTB dataset (model\# $4,5,$ and $6$). Furthermore, in designing these architectures we controlled for the effect of model size on performance by ensuring that for each of the $3$ models whose dilations were fitted to the dependency decay curve there was an equivalent sized model in the set of models using the standard pattern of dilations. Following the analysis of the PTB dataset we observe the inflection point of the broken power-law decay to be ${\approx}12$. We designed one model (model\# $4$) with a dense set of dilations up to the inflection point and two using patterns of dilations that follow the broken power-law pattern. Of these two, one model (model\# $5$) used standard set of dilations up to the inflection point and sparse dilations beyond and the second model (model\# $6$) used the dilation patterns that we believe most closely fits the broken power-law dependence decay of PTB dataset. The results of these six models are listed in table~\ref{tab:rnn_per}. 

One finding from these experiments is that larger models do not always outperform smaller models, for example models $1$ and $4$ outperform models $2,3,$ and $5$. More relevantly, however, if we compare models of the same size but with different patterns of dilations i.e. comparing model $1$ with $4$, model $2$ with $5$, and model $3$ with $6$, we find that in each case the model that uses pattern of dilations suggested by our framework (models $4,5,$ and $6$) outperforms the corresponding standard dilations model (models $1,2,$ and $3$). Furthermore, the best overall performance is achieved by model $6$ which uses a broken-power law pattern of dilations. The custom set of dilations used by model $6$ maximizes it's memory capacity over the steeper region of the dependency decay curve before the inflection point by introducing dense dilations. However, beyond the inflection point the dilations are sparse allowing for longer dependencies to be modeled by using less memory capacity. We interpret these results as confirming that best model performance is obtained when the pattern of dilations is customized to fit the dependency decay curve of the dataset being modeled.

\section{Discussion \& Conclusion}\label{sec:discussion}
In this paper. we have set out an approach to the design the hyper-parameter search space for recurrent neural architectures. The basis of this work is to use mutual information to analyse the dependency decay of a dataset. The dependency decay curve of a dataset is indicative of the presence of a specific dependency decay pattern in the dataset. For example, our analysis of sequential MNIST and character-based datasets indicate that the dependency decay of sequential MNIST is very different from character-based datasets. Understanding the properties of the underlying grammar that produces a sequence can aid in designing the grid-search space for the hyper-parameters to model a given dataset. In this work, we have used DilatedRNNs to illustrate our approach, and have demonstrated how customising the dilations to fit the dependency decay curve of a dataset improves the performance of the DilatedRNNs. However, we believe that the core idea of this approach can be more generally applied. Different neural architectures use different mechanisms to model LDDs, however these mechanisms have hyper-parameters that control how they function and we argue that it is useful to explicitly analyse the dependency decay curve of a dataset when selecting these hyper-parameters for the network.

\section*{Acknowledgments}
The research is supported by TU Dublin Scholarship Award. This research is also partly supported by the ADAPT Research Centre, funded under the SFI Research Centres Programme (Grant 13/ RC/2106) and is co-funded under the European Regional Development Funds. We also gratefully acknowledge the support of NVIDIA Corporation with the donation of the Titan Xp GPU under NVIDIA GPU Grant used for this research.

\bibliographystyle{splncs04}
\bibliography{ldds}

\end{document}